\documentclass[twoside,leqno,twocolumn]{article}
\usepackage{ltexpprt}
\usepackage{amsmath}
\usepackage{amsfonts}
\usepackage{mathabx}

\newcommand{\graph}{\mathcal{G}}
\newcommand{\vertices}{V}
\newcommand{\adjacency}{A}

\newcommand{\degrees}{D}
\newcommand{\laplacian}{L}
\newcommand{\vertex}{N}
\newcommand{\linear}{g}

\newcommand{\networkInput}{x}
\newcommand{\signal}{x}
\newcommand{\batch}{b}
\newcommand{\channel}{p}
\newcommand{\inputNeuron}{i}
\newcommand{\feature}{q}
\newcommand{\outputNeuron}{j}
\newcommand{\coef}{w}
\newcommand{\weightSharing}{s}
\newcommand{\parameters}{\theta}
\newcommand{\parameter}{k}
\newcommand{\matrices}[1]{\text{$\mathbf{\MakeUppercase{#1}}$}}
\newcommand{\vectors}[1]{\text{$\mathbf{#1}$}}
\newcommand{\maj}[1]{\text{\MakeUppercase{#1}}}

\begin{document}

\title{\Large A unified deep learning formalism for processing graph signals}
\author{Myriam Bontonou\thanks{IMT Atlantique, MILA} \\
\and
Carlos Lassance\footnotemark[1] \\
\and
Jean-Charles Vialatte\thanks{IMT Atlantique} \\
\and
Vincent Gripon\footnotemark[1]}
\date{}

\maketitle

\section{Introduction}
Deep learning networks outperform any other algorithms in many machine learning challenges. These networks rely on an assembly of layers. Each layer composes a linear function, whose coefficients are parameters to be learned, and a non linear function. The network is trained end-to-end, using a variant of gradient descent on a criterion to be optimized.
In images, a major factor explaining their performances is the ability of a layer to learn which shape features should be extracted in images. In such layers, a convolutional operator is implemented as a learnable linear function.

Two features distinguish the convolutional layer from a fully connected layer: a- the same parameters are used on all local portions of the input, b- the outputs of the network depend only on a local portion of the input. From these two characteristics arises an essential property of the convolutional layers: the number of parameters is independent of the size of the input.

This convolutional layer only makes sense when it is applied on inputs supported on a discrete Euclidean space (e.g: images, sounds). However, so many real-world signals are defined on more complex topologies (point clouds, geo-localized sensor networks, or functional activity in the brain). Thus, in recent years, a variety of works have emerged to extend the performances of convolutional layers to more diverse signals, and in particular, to signals defined on graphs.

In this study, we review some of the major deep learning models designed to exploit the underlying graph structure of signals. We express them in a unified formalism, giving them a new and comparative reading.

\section{Unified formalism}
Many deep learning approaches have been proposed to process graph signals. Each approach proposes a new linear function. As authors introduce their own formalism, it can be difficult to compare models. We propose a unified formalism to express them. Indices are written in lower case, and the size of their sets in upper case. In deep learning, data is divided into $\maj{\batch}$ batches. 
Given $\maj{\inputNeuron}$ input neurons and $\maj{\channel}$ channels, the input of a layer is a tensor $\matrices{\networkInput} \in \mathbb{R}
^{\maj{\batch} \times 
\maj{\channel} \times 
\maj{\inputNeuron}}$.
Similarly, given $\maj{\outputNeuron}$ output neurons and $\maj{\feature}$ features, the output $\linear(\matrices{\networkInput})\in\mathbb{R}^{\maj{\batch}\times \maj{\feature}\times \maj{\outputNeuron}}$.
Let's introduce a first simple example. Considering only one batch, $\matrices{\networkInput}\in\mathbb{R}^{\maj{\channel} \times \maj{\inputNeuron}}$, a feature $\feature$ and an output neuron $\outputNeuron$, we store the coefficients of the linear function $\linear$ in $\matrices{\coef}\in\mathbb{R}^{\maj{\feature} \times \maj{\channel} \times \maj{\outputNeuron} \times \maj{\inputNeuron}}$:
\begin{equation}
\linear(\matrices{\networkInput})_{\feature \outputNeuron} = 
\sum_{\channel=1}^{\maj{\channel}}
\sum_{\inputNeuron=1}^{\maj{\inputNeuron}}
\matrices{\coef}_{\feature \channel \outputNeuron \inputNeuron} \matrices{\networkInput}_{\channel \inputNeuron}\;.
\end{equation}

The same coefficients may be used in multiple connections. It is therefore inefficient to store them all in $\matrices{\coef}$. Subsequently, given $\maj{\parameter}$ unique parameters, we propose to store them in $\vectors{\parameters} \in \mathbb{R} ^ {\maj{\parameter}}$.
Now, we define an allocation tensor
$\matrices{\weightSharing}\in\mathbb{R}^{\maj{\parameter} \times \maj{\outputNeuron} \times \maj{\inputNeuron}}$
such that $\matrices{\coef}_{\feature \channel \outputNeuron \inputNeuron} = 
\sum_{\parameter=1}^{\maj{\parameter}}
\vectors{\parameters}_{\feature \channel \parameter} \matrices{\weightSharing}_{\parameter\outputNeuron\inputNeuron}$.
To simplify the notations, we omit the sums and by convention, we consider that there is a sum on an index when it is no longer present in the output variable. The formula above becomes:
$\matrices{\coef}_{\feature \channel \outputNeuron \inputNeuron} = \vectors{\parameters}_{\feature \channel \parameter} \matrices{\weightSharing}_{\parameter \outputNeuron \inputNeuron}$.
By reintroducing the batch index, we obtain the following formalism:
\begin{equation}
    \label{eq1}
    \linear(\matrices{\networkInput}) = 
    \widehat{ \matrices{\Theta} \matrices{\weightSharing} \matrices{\networkInput} }
    \;\text{where}\; 
\left \{
  \begin{tabular}{ccc}
  $\matrices{\coef}_{\feature \channel \outputNeuron \inputNeuron} = 
  \matrices{\Theta}_{\feature \channel \parameter} \matrices{\weightSharing}_{\parameter \outputNeuron \inputNeuron}$  \\
  $\linear(\matrices{\networkInput})_{\batch \feature \outputNeuron} = \matrices{\coef}_{\feature \channel \outputNeuron \inputNeuron}
  \matrices{\networkInput}_{\batch \channel \inputNeuron}$ \\
  \end{tabular}
\right \}
\end{equation}

\section{Results and Discussion}
The formalism  $\linear(\matrices{\networkInput}) = \widehat{\matrices{\Theta} \matrices{\weightSharing} \matrices{\networkInput}}$ distinguishes the roles of $\matrices{\Theta}$ and $\matrices{\weightSharing}$. $\matrices{\Theta}$ stores the parameters. $\matrices{\weightSharing}$ contains the parameter sharing scheme. In this section, we express four major deep learning models in graph signal processing as well as a fully connected (FC) layer and a conventional convolutional layer. Consider a graph $\graph = \langle \vertices, \matrices{\adjacency} \rangle$, where $\vertices = \{1\dots\vertex\}$ is a set of vertices, $\matrices{\adjacency}\in\mathbb{R}^{\vertex\times \vertex}$ an adjacency matrix, $\matrices{\degrees}$ a degree matrix and $\matrices{\laplacian} = \matrices{\degrees} - \matrices{\adjacency}$ a Laplacian matrix. A signal with $\maj{\channel}$ features, supported on a graph, is represented by $\matrices{\signal}\in\mathbb{R}^{\vertex\times \maj{\channel}}$. This signal can be an input of the linear function described in the unified formalism.

{\noindent}\textbf{Fully connected layer}: A FC layer is represented by $\maj{\inputNeuron}$ input neurons, $\maj{\outputNeuron}$ output neurons, 1 channel and 1 feature. The number of parameters $\maj{\parameter}$ is equal to the number of possible connections between the input neurons and the output neurons, $\maj{\parameter} = \maj{\inputNeuron}\times \maj{\outputNeuron}$. In that case, $g(\matrices{\networkInput})_{b1\outputNeuron} = \matrices{\Theta}_{11k}\matrices{\weightSharing}_{\parameter\outputNeuron\inputNeuron}\matrices{\networkInput}_{b1\inputNeuron}$. $\matrices{\Theta}_{11k}\in\mathbb{R}^{1 \times 1 \times \maj{\inputNeuron}\maj{\outputNeuron}}$ is indeed a vector. Given neurons $\inputNeuron$ and $\outputNeuron$, $\matrices{\weightSharing}[:,\outputNeuron,\inputNeuron]$ is a one-hot vector, which selects the parameter associated to the connection between $\inputNeuron$ and $\outputNeuron$. By squeezing the shape of tensors, we get the usual formula describing a FC layer: $g(\matrices{\networkInput})_{\batch\outputNeuron} = \matrices{\coef}_{\outputNeuron\inputNeuron}
  \matrices{\networkInput}_{\batch\inputNeuron}$.

{\noindent}\textbf{Convolutional layer}: Parameters depend on the inputs channel $\channel$ and feature map $\feature$. $\matrices{\Theta}[\feature,\channel,:]\in\mathbb{R}^{\maj{\parameter}}$ contains the kernel weights for a given input channel $\channel$ and feature map $\feature$. In a convolutional layer, the convolution matrix is Toeplitz. Each diagonal, from the top left to the bottom right, contains either the same parameter or a zero. In the ternary representation, $\matrices{\weightSharing}$ maintains the same structure. Indeed, in the 1D case,  $\matrices{\weightSharing}[\parameter,:,:]\in\mathbb{R}^{\maj{\inputNeuron} \times \maj{\outputNeuron}}$ is full of zeros, except on one diagonal where the coefficient is 1. For each parameter $\matrices{\Theta}[\feature,\channel,\parameter]$, a different diagonal of $\matrices{\weightSharing}$ would contain 1.

{\noindent}\textbf{ChebNet} ChebNet~\cite{defferrard2016convolutional} is inspired from spectral graph theory, where the Discrete Fourier Transform (DFT) is expanded to graph signals. Let $c$ be the order of a Chebyshev polynomial $T_c$. The linear function $\linear$ becomes:
\begin{equation}
\linear(\matrices{\signal})_{\feature} = \sum_{\channel=1}^{\maj{\channel}}\matrices{\coef}_{\feature\channel}(\matrices{\laplacian})\vectors{\signal}_{\channel}\;.
\end{equation}
Given a constant $\maj{c}$, $\lambda_{\max}$ the largest eigenvalue of $\matrices{\laplacian}$, the identity matrix $\mathbf{I}_\vertex$, $T_c(\frac
{\lambda_{\max}\matrices{\laplacian}}
{2} - \mathbf{I}_\vertex)\in\mathbb{R}^{\vertex\times \vertex}$ and $\vectors{\parameters}\in\mathbb{R}^{\maj{c}}$ the learnable parameters, $\matrices{\coef}_{\feature\channel}\in\mathbb{R}^{\vertex \times \vertex}$ is defined as:
$\matrices{\coef}_{\feature\channel}(\matrices{\laplacian}) = \sum_{c=0}^{\maj{c}-1}\vectors{\parameters}_{c}\;T_c(\frac
{\lambda_{\max}\matrices{\laplacian}}
{2} - \mathbf{I}_\vertex)\;.$

\vspace{0.1cm}
\underline{Unified formalism:} There are as many input neurons $\maj{\inputNeuron}$ as output neurons $\maj{\outputNeuron}$ as vertices in the graph $\maj{\inputNeuron}=\maj{\outputNeuron}=\vertex$. Only one weight is computed for a couple ($\channel$,$\feature$), therefore $\maj{\parameter}=1$. When $\matrices{\weightSharing}=\sum_{c=0}^{C-1}T_c(
\frac
{\lambda_{\max}(\matrices{\laplacian})\matrices{\laplacian}}
{2} - \mathbf{I}_\vertex)$, a Chebyshev filter is computed.

{\noindent}\textbf{Graph Convolutional Network}
GCN~\cite{kipf2016semi} is a neural network using the following linear function $\linear$:
\begin{equation}
    g(\matrices{\signal}) = \Tilde{\matrices{\adjacency}}\;\matrices{\signal}\;\matrices{\Theta}\;,
\end{equation}
where $\Tilde{\matrices{\adjacency}} = \hat{\matrices{\degrees}}^{-1/2}\hat{\matrices{\adjacency}}\hat{\matrices{\degrees}}^{-1/2}$, $\hat{\matrices{\adjacency}} = \matrices{\adjacency} + \matrices{I}_{N}$, $\hat{\matrices{\degrees}}$ degree matrix of $\hat{\matrices{\adjacency}}$, $\matrices{\Theta}\in\mathbb{R}^{\maj{\channel} \times \maj{\feature}}$.
$\Tilde{\matrices{\adjacency}}\matrices{\signal}$ diffuses the signal on the graph. Now, the values of the signal on a vertex depend on the values of the neighboring vertex signals. $\matrices{\Theta}$ learns several representations of the diffused signal.
This model is close to ChebNet if we reformulate it as follows:
$g(\matrices{\signal}) = \sum_{c=0}^{C-1}T_c\left(\frac
{\lambda_{\max}\matrices{\laplacian}}
{2} - \mathbf{I}_\vertex\right)\;\matrices{\signal}\;\matrices{\Theta}\;.$

\vspace{0.1cm}
\underline{Unified formalism:} Consequently, its expression is similar as ChebNet. $\maj{\inputNeuron}=\maj{\outputNeuron}=\vertex$. $\maj{\parameter}=1$. $\matrices{\weightSharing}$ corresponds to the normalized adjacency matrix.

{\noindent}\textbf{Graph ATtention network} 
GAT~\cite{velivckovic2017graph} learns parameters which measure the importance of the neighbors considering all vertices independently. Given a matrix $\matrices{\Theta}\in\mathbb{R}^{\maj{\channel}\times \maj{\feature}}$ and two neighbor vertices $(i,j)$, a fully connected layer, represented by the vector $\mathbf{a}\in\mathbb{R}^{2\maj{\feature}}$, learns the attention $\alpha_{ji}$ that $j$ deserves from $i$. $\alpha_{ji} = softmax(e_{ji})$, where $e_{ji} = LeakyReLU(\mathbf{a}^{T}\;[\matrices{\signal}_j\matrices{\Theta}\;||\;\matrices{\signal}_i\matrices{\Theta}])$, $||$ meaning concatenation.
$\matrices{\Theta}$ transforms input features into higher-level features, and $\mathbf{a}$ diffuses the signal. To exploit the structure of the graph, a mechanism called attention mask is set up: $e_{ji}$ is zero if the vertex $i$  is not connected to the vertex $j$. 
Finally, noting $\mathcal{R}(j)$ the set of vertices connected to the vertex $j$, we get the following linear function $\linear$:
\begin{equation}
    g(\matrices{\signal})_{j} = \sum_{i\in\mathcal{R}(j)}
    \alpha_{ji}\matrices{\signal}_i\matrices{\Theta}\;.
\end{equation}
To stabilize learning, several linear functions $\linear$ are computed independently, then concatenated or averaged.

\vspace{0.1cm}
\underline{Unified formalism:} Similarly to GCN, $\maj{\inputNeuron}=\maj{\outputNeuron}=\vertex$. To express a concatenated multiple attention heads layer, $A$ ternary layers are concatenated. For each layer, as only one weight is computed for a couple ($\channel$,$\feature$), $\maj{\parameter}=1$, and $\matrices{\weightSharing}$ contains the attention coefficients. To express an averaged multiple attention heads layer, $A$ weights are computed for a couple ($\channel$,$\feature$), $\maj{\parameter}=A$, and $\matrices{\weightSharing}[\parameter,:,:]$ contains the attention coefficients divided by $A$.

{\noindent}\textbf{Topology Adaptative GCN}
TAGCN~\cite{du2017topology} makes the signal of a vertex dependent to the signals of its neighbors located at most $C$ connections. It introduces the powers of the normalized adjacency matrix $\Tilde{\matrices{\adjacency}}$ in GCN:
\begin{equation}
    g(\matrices{\signal}) = \sum_{c=1}^{C}\Tilde{\matrices{\adjacency}}^c\;\matrices{\signal}\;\matrices{\Theta}_c\;.
\end{equation}

\vspace{0.1cm}
\underline{Unified formalism:} Similar to GCN, excepted that the weight sharing $\matrices{\weightSharing}[\parameter,:,:]$ contains the normalized adjacency matrix $\Tilde{\matrices{\adjacency}}^\parameter$. Plus, if at most $C$ powers are computed, given a couple ($\channel$,$\feature$), $C$ parameters have to be learned. So, $\maj{\parameter}=C$.

\bibliographystyle{siamplain}
\bibliography{ltexpprt}

\end{document}